\documentclass[letterpaper, 10 pt, conference]{ieeeconf}  

\IEEEoverridecommandlockouts                              

\usepackage{amsmath} 
\usepackage{amssymb}  
\usepackage{graphicx}

\usepackage{algorithm}
\usepackage{algpseudocode}
\newcommand{\algorithmicbreak}{\textbf{break}}
\newcommand{\Break}{\State \algorithmicbreak}
\usepackage[hidelinks]{hyperref}
\usepackage{tcolorbox}

\title{\LARGE \bf
Generating and Evolving Reward Functions for Highway Driving with Large Language Models
}

\author{Xu Han$^{1\dag}$, Qiannan Yang$^{2\dag}$, Xianda Chen$^{2}$, Xiaowen Chu$^{1}$, Meixin Zhu$^{2*}$
\thanks{This study is partly supported by the National Natural Science Foundation of China under Grant 52302379, Guangzhou Basic and Applied Basic Research Project 2023A03J0106, Guangdong Province General Universities Youth Innovative Talents Project under Grant 2023KQNCX100, and Guangzhou Municipal Science and Technology Project 2023A03J0011.}
\thanks{$^{1}$Xu Han, Xiaowen Chu are with the Data Science and Analytics Thrust, Information Hub, The Hong Kong University of Science and Technology (Guangzhou), Guangzhou, China
        {\tt\small xhanab@connect.ust.hk, xwchu@ust.hk}}%
\thanks{$^{2}$Qiannan Yang, Xianda Chen, Meixin Zhu are with the Intelligent Transportation Thrust, Systems Hub, The Hong Kong University of Science and Technology (Guangzhou), Guangzhou, China; also with Guangdong Provincial Key Lab of Integrated Communication, Sensing and Computation for Ubiquitous Internet of Things
        {\tt\small {qyangan, xchen595}@connect.hkust-gz.edu.cn, meixin@ust.hk}}%
\thanks{\dag Xu Han and Qiannan Yang contributed equally to this work.}%
\thanks{*Corresponding author: Meixin Zhu.
        }%
}

\begin{document}

\maketitle
\thispagestyle{empty}
\pagestyle{empty}

\begin{abstract}

Reinforcement Learning (RL) plays a crucial role in advancing autonomous driving technologies by maximizing reward functions to achieve the optimal policy. However, crafting these reward functions has been a complex, manual process in many practices. To reduce this complexity, we introduce a novel framework that integrates Large Language Models (LLMs) with RL to improve reward function design in autonomous driving. This framework utilizes the coding capabilities of LLMs, proven in other areas, to generate and evolve reward functions for highway scenarios. The framework starts with instructing LLMs to create an initial reward function code based on the driving environment and task descriptions. This code is then refined through iterative cycles involving RL training and LLMs' reflection, which benefits from their ability to review and improve the output. We have also developed a specific prompt template to improve LLMs' understanding of complex driving simulations, ensuring the generation of effective and error-free code. Our experiments in a highway driving simulator across three traffic configurations show that our method surpasses expert handcrafted reward functions, achieving a 22\% higher average success rate. This not only indicates safer driving but also suggests significant gains in development productivity.


\end{abstract}

\section{INTRODUCTION}

\begin{figure}
\centering
\includegraphics[width=3.4in]{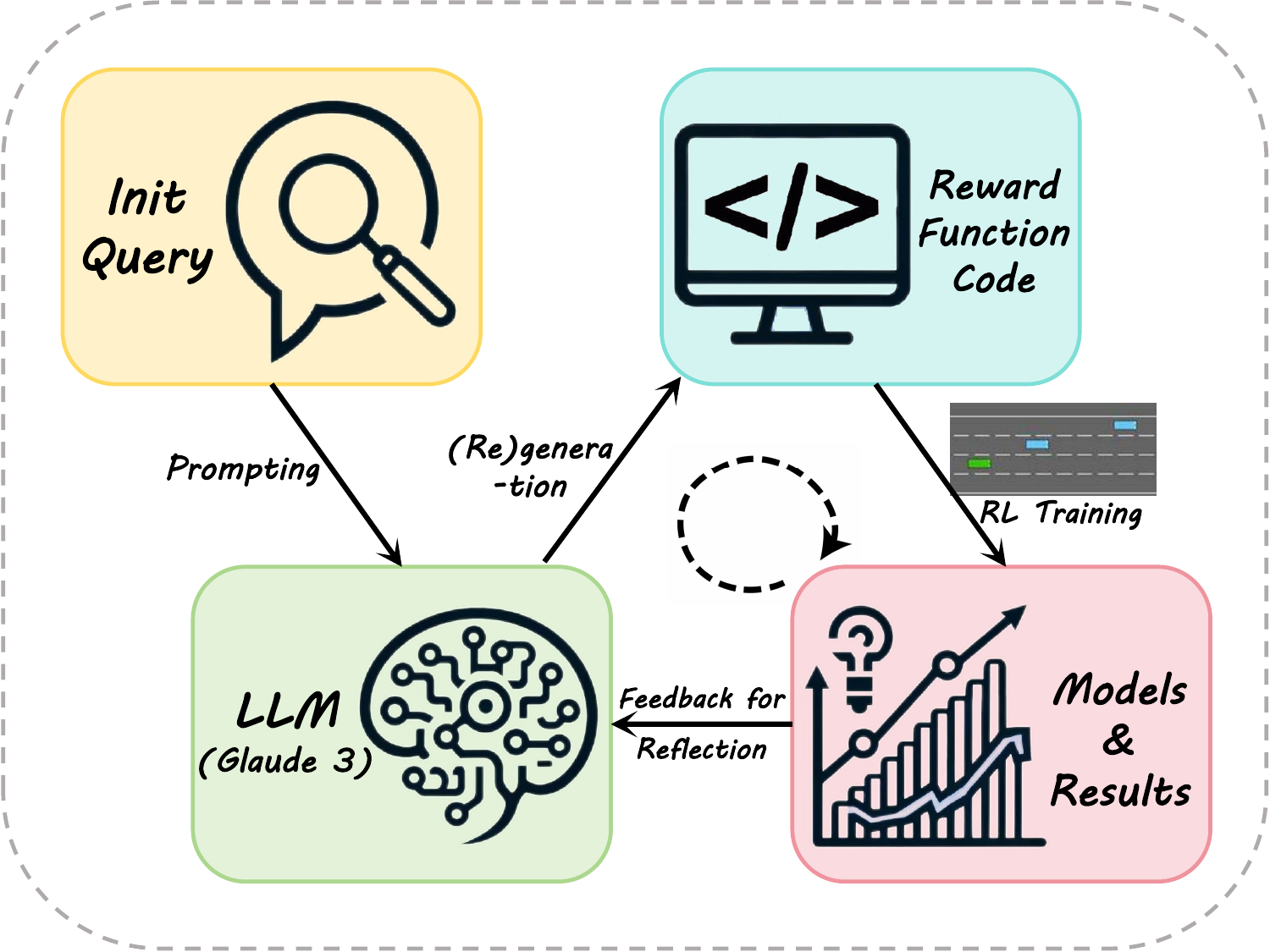}
\caption{Conceptual diagram of the proposed framework. LLMs generate reward function codes for driving according to user instructions by using an elaborate prompt template. Then the results of RL training based on the designed reward are fed back to LLMs for reflection and reward regeneration, aiming for evolutionary improvements.}
\label{Figure 1}
\end{figure}

Large Language Models (LLMs) have shown remarkable abilities and are now being explored in autonomous driving, primarily for tasks such as semantic analysis, logical reasoning, and decision-making \cite{wen2023dilu, nouri2024engineering, cui2024drive, dewangan2023talk2bev}. Despite this, only a few studies have investigated the potential of LLMs in coding for autonomous driving applications \cite{ma2023lampilot, ishida2024langprop}.

Reinforcement learning (RL), on the other hand, has been a staple in autonomous driving research, focusing on how vehicles can autonomously navigate complex traffic scenarios through exploration and exploitation in simulations \cite{sallab2017deep, cao2022trustworthy}. The design of the reward functions in RL research with driving tasks, crucial for the learning process, remains a significant challenge due to its reliance on the manual, often tedious trial-and-error process \cite{booth2023perils, knox2023reward, chen2023follownet}.

Recognizing the gap in efficiently designing reward systems in autonomous driving, we propose leveraging LLMs to innovate this process. Inspired by the use of coding LLMs in other fields \cite{chen2024evoprompting, ma2023eureka, guo2023connecting}, our framework aims to generate and evolve reward functions for highway driving tasks. This approach is grounded in LLMs’ demonstrated strengths in understanding human driving behaviors and coding ability, offering a novel way to potentially streamline the reward design process \cite{cui2024survey}.

Our framework introduces an iterative process, as shown in Fig. \ref{Figure 1}, where LLMs generate reward function code according to driving environment code and task description, which are then refined through RL training, feedback and reflection, and regeneration, aiming for evolutionary improvements. This method is motivated by the unique challenges in autonomous driving, including the complexity of simulating realistic driving environments, the dynamic nature of traffic scenarios, and the great emphasis on driving safety \cite{yan2023learning}. These challenges necessitate sophisticated reward function designs that can adapt to a wide range of driving conditions and behaviors.

To address the intricate simulation code structures and dynamic driving conditions, we designed an elaborate prompt template based on the task characteristics of highway driving, which significantly improved LLMs' comprehension of the highway driving environment and led to the generation of high-quality, error-free reward function code, marking a novel application of LLMs' coding capabilities in this area. The contributions of this work to the autonomous driving research field include:

\begin{itemize}
\item Introducing a framework that utilizes LLMs for the generation and evolution of reward functions in driving tasks, demonstrating the potential to alleviate human workload and enhance productivity.
\item Implementing carefully designed prompt templates to enhance LLMs' understanding of complex driving simulation codes, facilitating the generation of effective and error-free reward function codes.
\item Demonstrating through extensive testing with a highway driving simulator that our method can outperform expert human-designed rewards across various traffic conditions, achieving an average success rate improvement of 22\%, which identifies the great potential for safe and efficient autonomous driving.
\end{itemize}

The remainder of this paper is structured as follows. Section \ref{section:related} reviews relevant prior work. Section \ref{section:approach} details the proposed framework. Section \ref{section:expe} introduces the experiment design and analyzes the efficacy of the proposed approach. Section \ref{section:conclusion} draws the conclusions.

\section{RELATED WORKS} \label{section:related}

\subsection{Large Language Models for Autonomous Driving}

Recent advancements in autonomous driving technology have been increasingly influenced by the integration of Large Language Models (LLMs), marking a significant shift towards more intelligent and adaptable systems. These models have been employed across various aspects of autonomous driving, from enhancing decision-making processes to improving simulation frameworks and ensuring safety through advanced requirement management.

The DiLu framework \cite{wen2023dilu} embodies a pioneering approach by integrating reasoning and reflection capabilities, showcasing significant advancements in system adaptability and real-world application readiness. Meanwhile, Surrealdriver \cite{jin2023surrealdriver} leverages generative simulation to reduce collision rates and enhance the realism of driver behaviors in urban settings. Advancements are not confined to simulation and decision-making; the integration of LLMs for engineering safety requirements demonstrates their critical role in refining safety protocols, ensuring the dynamic automotive domain remains secure and reliable \cite{nouri2024engineering}. The exploration of text-based and multimodal inputs for traffic scene representation and decision-making illustrates the breadth of LLM application, significantly improving scene understanding and prediction accuracy \cite{keysan2023can}. Frameworks like those proposed in \cite{cui2024drive} for human-like interaction within autonomous vehicles aim to revolutionize passenger experience by offering personalized assistance and seamless decision-making. Similarly, innovations such as Talk2BEV \cite{dewangan2023talk2bev} and LanguageMPC \cite{sha2023languagempc} showcase the potential of LLMs in enhancing visual reasoning and commonsense decision-making in driving scenarios. On the cutting edge, GAIA-1 \cite{hu2023gaia} introduces a generative world model that predicts complex driving scenarios, underscoring the significance of unsupervised learning in autonomous driving. Projects like ChatGPT As Your Vehicle Co-Pilot \cite{wang2023chatgpt} and TrafficGPT \cite{zhang2024trafficgpt} illustrate the practical applications of LLMs in improving the synergy between human intentions and machine executions, advancing urban traffic management through insightful AI-driven solutions. DriveCoT \cite{wang2024drivecot} and VLAAD \cite{park2024vlaad} focus on enhancing interpretability and controllability in driving decisions, employing LLMs for better navigation and instruction comprehension. 

Moreover, LaMPilot \cite{ma2023lampilot}, LangProp \cite{ishida2024langprop}, and ChatSim \cite{wei2024collaborative} utilizes the coding ability of LLMs, introducing novel frameworks for code optimization and editable scene simulation, which highlights the importance of LLMs in achieving transparent and adaptable autonomous driving solutions.

Collectively, these advancements underscore the vital role of LLMs in pushing the boundaries of autonomous driving technology, offering novel solutions for safety, efficiency, and user experience. For a more comprehensive review of LLMs for autonomous driving, the works of \cite{yang2023llm4drive} and \cite{cui2024survey} are recommended.

\subsection{Deep Reinforcement Learning}

Deep Reinforcement Learning (DRL) is an advanced field combining the strengths of RL and deep learning (DL), enabling agents to learn and make decisions in complex environments \cite{mnih2015human}. A common formulation of the DRL problem is the Markov Decision Process (MDP), a mathematical framework that models decision-making in situations where outcomes are partly random and partly under the control of a decision-maker. MDPs are characterized by states (\(s\)), actions (\(a\)), transition probabilities (\(P(s'|s, a)\)), and rewards (\(R(s, a)\)), offering a systematic way to describe the dynamics of an environment.

The Bellman equation \cite{sutton1999reinforcement}, a fundamental component of MDPs, provides a recursive relationship essential for understanding and solving reinforcement learning problems. It is expressed as:

\begin{small}
\begin{equation}
\label{eqn:1}
Q^*(s, a) = \mathbb{E}[R_{t+1} + \gamma \max_{a'}Q^*(s', a')].
\end{equation}
\end{small}

\noindent where \(Q^*(s, a)\) represents the optimal action-value function, indicating the expected return for taking action \(a\) in state \(s\) and following the best strategy afterward. \(R_{t+1}\) is the immediate reward received, and \(\gamma\) is the discount factor, which quantifies the importance of future rewards. The variables \(s'\) and \(a'\) denote the subsequent state and action, respectively. The expectation \(\mathbb{E}[\cdot]\) accounts for the stochastic nature of the environment. By iteratively applying this equation, DRL algorithms aim to approximate \(Q^*\), guiding agents towards maximizing their cumulative rewards, or the so-called optimal policy.

Many innovative applications of RL in the realm of autonomous driving have demonstrated their potential to address complex, dynamic, and uncertain environments \cite{sallab2017deep, lu2023event}. Through various frameworks and simulations, RL is shown to effectively teach machines to navigate and make decisions like human drivers, achieving human-like car-following behaviors, and handling diverse driving conditions with improved accuracy \cite{osinski2020simulation, zhu2018human}. These studies underscore RL's capacity for continuous learning and adaptation, proposing hybrid models that combine RL with other methodologies for enhanced safety and performance \cite{han2023ensemblefollower}. Furthermore, the integration of RL with rule-based algorithms in a decision-making framework showcases a pathway towards achieving trustworthy and intelligent autonomous driving systems, capable of self-improvement and higher-level intelligence while ensuring safety \cite{cao2022trustworthy}.

\subsection{Reward Engineering for Autonomous Driving}

Reward engineering aims to solve the reward design problem \cite{singh2009rewards} in reinforcement learning, which involves creating a reward function that effectively guides an agent toward desired behaviors in an environment. This task is challenging due to the need to accurately represent complex objectives, prevent unintended behaviors, and ensure the agent's actions align with human standards and values. It is crucial for enabling the agent to learn efficiently and achieve optimal performance across a variety of scenarios or tasks, while also navigating issues such as sparse rewards, the balance between exploration and exploitation, and ensuring safety and robustness in real-world applications \cite{franccois2018introduction}.

Previous research has primarily used manual trial-and-error methods to address these issues, particularly in areas like autonomous driving, where the major goals are to align with human values and ensure robustness in different 

\begin{algorithm}
\caption{Generating and Evolving Reward Function}
\label{alg:1}
\begin{algorithmic}[1]
\Require Initial prompt $P_I$, reflection prompt $P_R$, environment code $EC$, Large Language Model $M$, evaluation function for reward $E$
\State Hyperparameters: number of iterations $N$, reward candidate size of each iteration $C$, evaluation threshold of reward $Q_{thres}$
\State // Generate $C$ initial reward candidates $\{ R \}$
\State $R_1, \ldots, R_C \sim M(P_I, EC)$
\For{$N$ iterations}
    \State // Obtain evaluation $Q$ for each reward candidate
    \State $Q_1 = E(R_1), \ldots, Q_C = E(R_C)$
    \State // Store the best-performing reward function and its evaluation
    \State $Q_{best} = \arg\max_{c} Q_1, \ldots, Q_C$
    \State $R_{best} = \arg_{Q_c = Q_{best}} R_1, \ldots, R_C$
    \State // Early stop if reaching threshold
    \If{$Q_{best} > Q_{thres}$}
        \Break
    \EndIf
    \State // Reflection and generate new reward candidates
    \State $R_1, \ldots, R_C \sim M(P_R, EC, R_{best}, Q_{best})$
\EndFor
\State \textbf{Output:} $R_{best}$
\end{algorithmic}
\end{algorithm}

\noindent driving conditions \cite{pan2019reinforcement, zhu2020safe, yuan2019multi, knox2023reward}. Although some work, such as \cite{chiang2019learning}, has explored automated reward search using evolutionary algorithms, these efforts have been limited to adjusting parameters within existing reward templates. Inverse reinforcement learning (IRL) offers a way to deduce reward functions from observed expert behavior \cite{rosbach2020driving, huang2021driving}. However, IRL depends on collecting high-quality expert data, which can be costly and is not always accessible. Moreover, it tends to produce rewards that are difficult to interpret.

In contrast, our approach can automatically generate and evolve reward functions. Unlike previous methods, our framework produces understandable reward function code without relying on human design or gradient calculations.

\section{PROPOSED APPROACH} \label{section:approach}

The proposed framework, designed to enhance the development and refinement of reward functions for driving simulation tasks, integrates LLMs with RL and a feedback loop for continuous improvement. For clarity, the pseudocode detailing our entire framework is available as Algorithm \ref{alg:1}, with supplementary prompt templates and guidelines for LLMs included in the appendix. The framework unfolds in three interconnected stages:

\subsection{Understanding Driving Simulation Environment}

\begin{figure}
\centering
\includegraphics[width=3.4in]{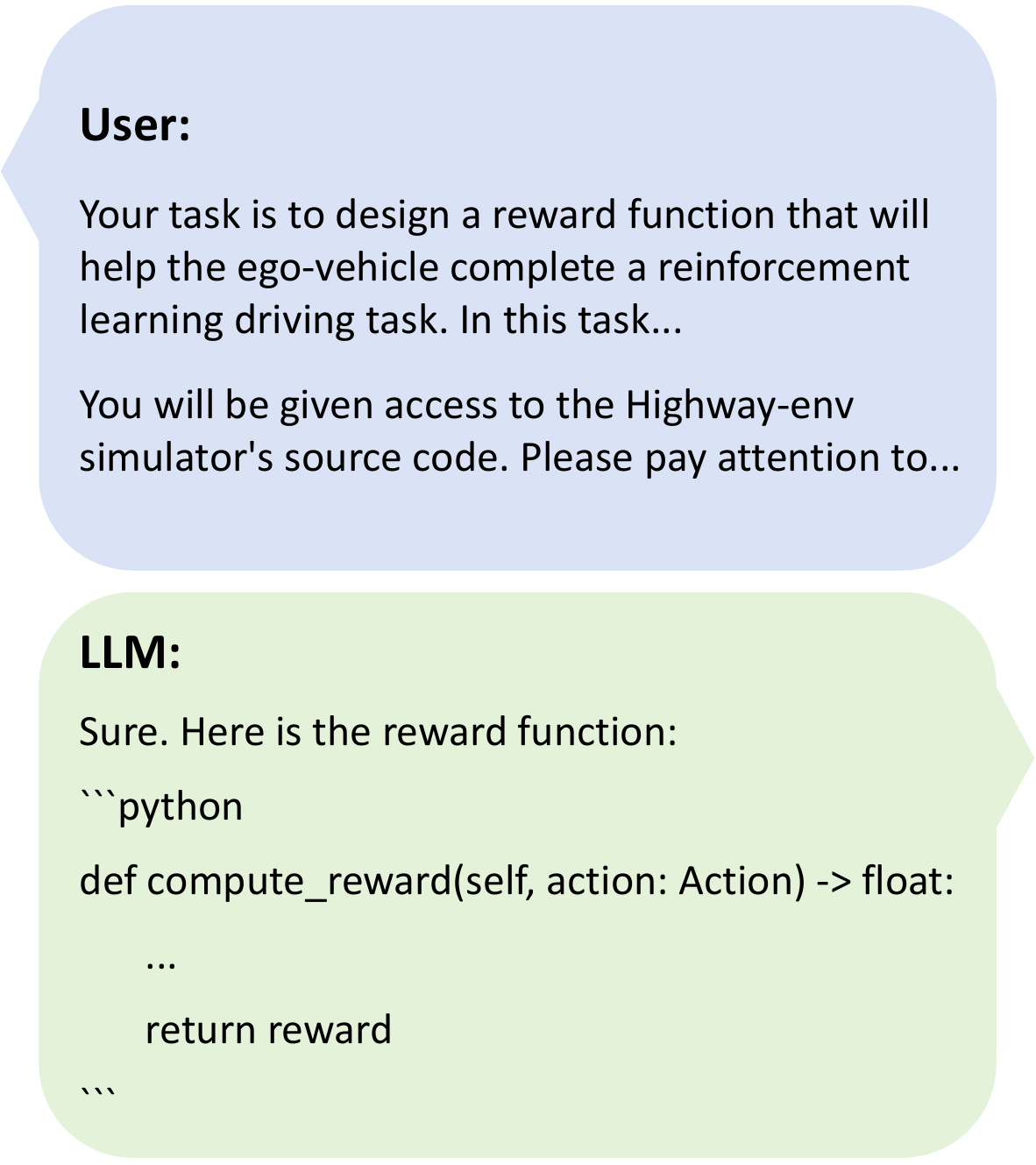}
\caption{Conversation example between user and LLM. The user prompt includes task description and environment source code, while LLM replies with a reward function.}
\label{Fig:rl_training}
\end{figure}

The foundation of our framework is enabling LLMs to grasp the nuances of the driving simulation environment. This is crucial for them to generate viable reward function codes. To accomplish this, we provide LLMs with detailed instructions, as shown in Fig. 2, which includes a task

\begin{figure*}[t!]
\centering
\includegraphics[width=0.83\textwidth]{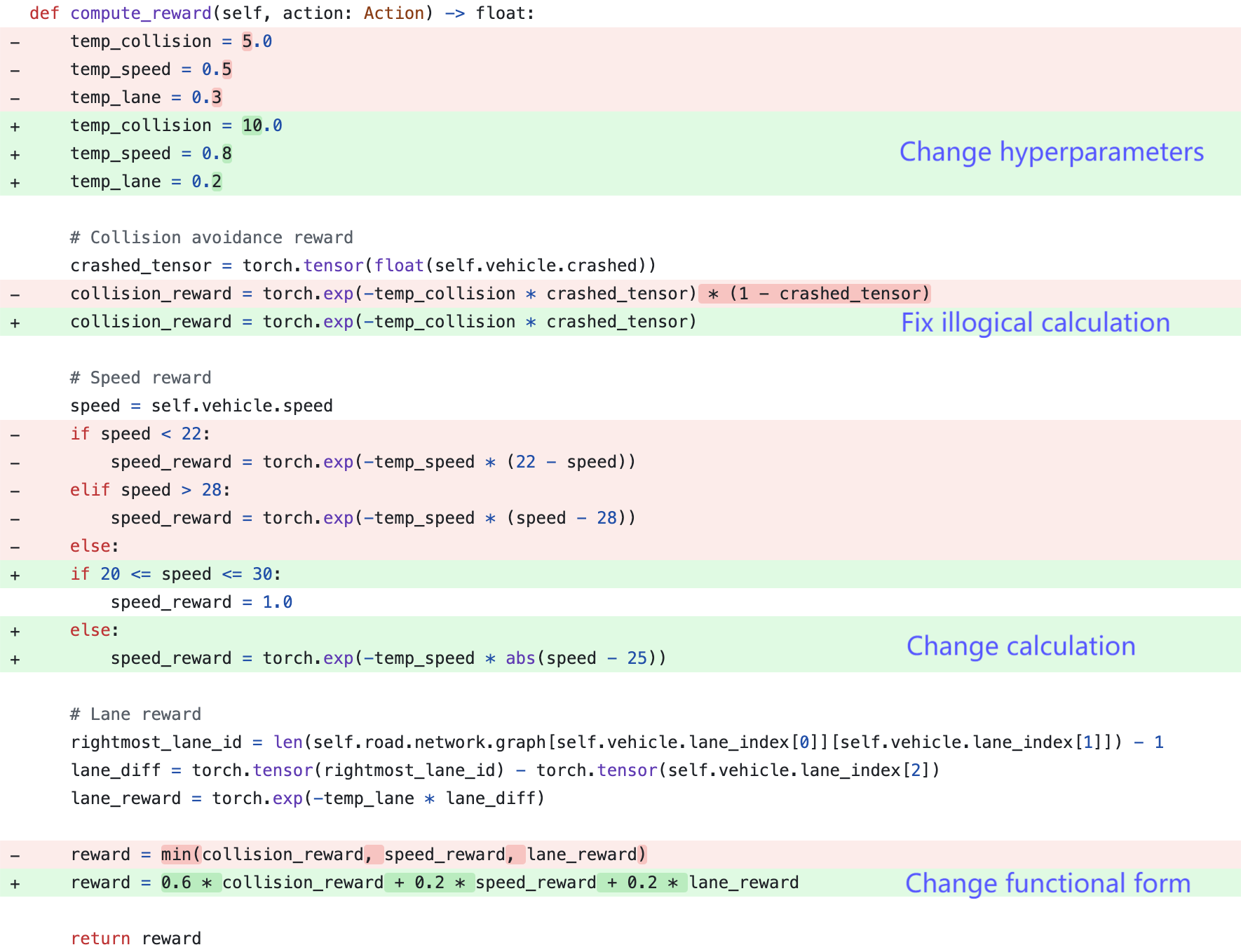}
\caption{Example of LLM refining the reward function in an iteration.}
\label{fig:reflection}
\end{figure*}

\noindent description (e.g., creating a reward function for safe, comfortable driving) and the simulation environment's code, excluding the reward function itself. The rationale is twofold: code offers a precise and concise medium for understanding the environment compared to natural language, and it allows LLMs to directly identify variables critical for reward function design. We designed an elaborate prompt template to guide LLMs through the environment's code systematically, addressing potential comprehension issues and ensuring the generated reward functions are executable and effective.

\subsection{Reinforcement Learning for Highway Driving}

Upon generating executable reward functions, we embed these into the simulation to conduct RL training for a highway driving agent. In each iteration, the proposed approach generates multiple independent samples of reward functions using LLMs, enhancing the search efficiency for robust reward functions and facilitating parallel training processes. This stage maintains consistent reinforcement learning algorithms and hyperparameters across iterations, either concluding upon reaching predefined objectives or after a set number of iterations.

\subsection{Reflection and Refinement}

Unlike traditional methods that rely on gradient descent for optimization, our framework seeks to refine reward functions through iterative feedback. After each RL training session, we describe the performance outcomes to the LLMs in natural language, providing them with specific metrics (like collision rates) and, potentially, more detailed feedback through arrays or tables to pinpoint areas for reward function improvement. This feedback prompts LLMs to propose modifications, ranging from comprehensive redesigns to targeted adjustments, enhancing the reward function's effectiveness and executability in subsequent iterations. Fig. \ref{fig:reflection} gives an example of how an LLM refines the reward function in an iteration.

This structure ensures that each stage logically flows into the next, from understanding the simulation environment and generating reward functions to applying these functions in RL and iteratively improving them based on performance feedback.

\section{EXPERIMENTS AND RESULTS} \label{section:expe}

\subsection{Experiment Design}

We evaluated our proposed framework in highway driving scenarios to assess its capability in generating effective reward functions and addressing new tasks. The initial experiments were conducted using GPT-4 (OpenAI, 2023) to design reward functions. However, subsequent tests showed that Claude 3-Haiku (Anthropic, 2024) was not only more effective but also more cost-efficient. Hence, Claude 3-Haiku became our primary LLM for all further experiments, unless otherwise specified. We chose the Highway-env platform (Leurent, 2018), which is known for its autonomous driving and tactical decision-making simulations. This platform features diverse driving models and realistic multi-vehicle interactions, allowing for variable vehicle density and lane configurations. These settings ensured that the LLM could not depend solely on pre-existing data, making it a robust environment for testing our framework's ability to generate new reward functions. The inputs of task description for Claude 3 were sourced directly from the official environment repository, and we compared the new rewards generated against the original ones—created by experts in reinforcement learning—which served as our baseline and are denoted as "Human" in our results.

\subsection{Training Details}

\begin{figure}
\centering
\includegraphics[width=3.4in]{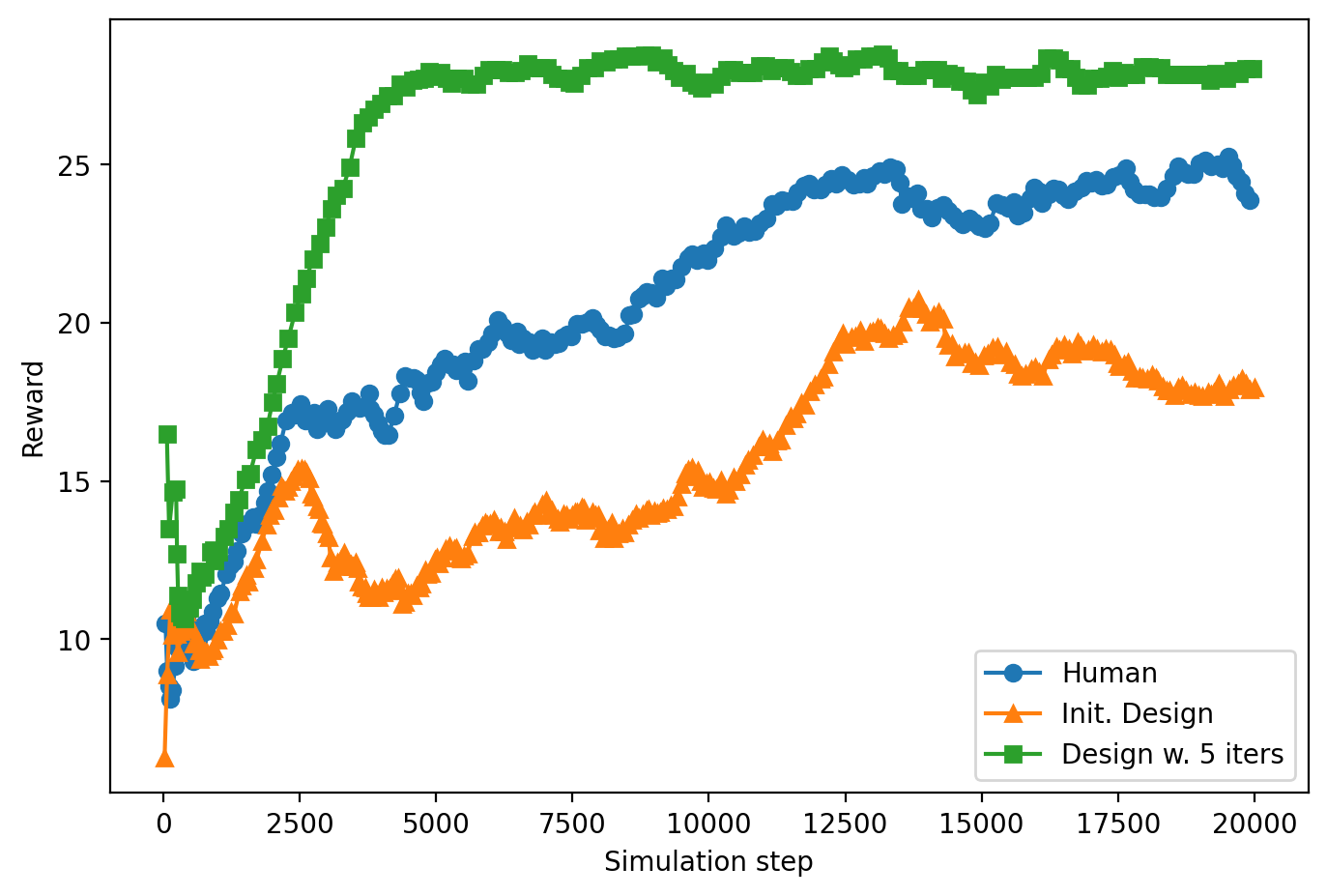}
\caption{Performance of generated and human rewards during RL training.}
\label{Fig:rl_training}
\end{figure}

Using the same DQN-based reinforcement learning setup provided by Highway-env, we optimized all reward functions across 20,000 training steps without altering the established hyperparameters. This setup had been pre-tuned to perform optimally with the official, manually designed rewards. Our observations indicated significant improvements in reward design through our framework, as depicted in Fig. \ref{Fig:rl_training}. Initially, the reward functions generated by our system underperformed compared to human-designed rewards. However, after five iterations, our system-generated rewards not only exceeded human performance but also demonstrated rapid convergence to the global optimum. The absolute scale of these three reward functions have subtle differences, which would not change our conclusion. This progression underscores our method's capacity for continual enhancement of reward design in driving tasks.

\begin{figure}
\centering
\includegraphics[width=3.4in]{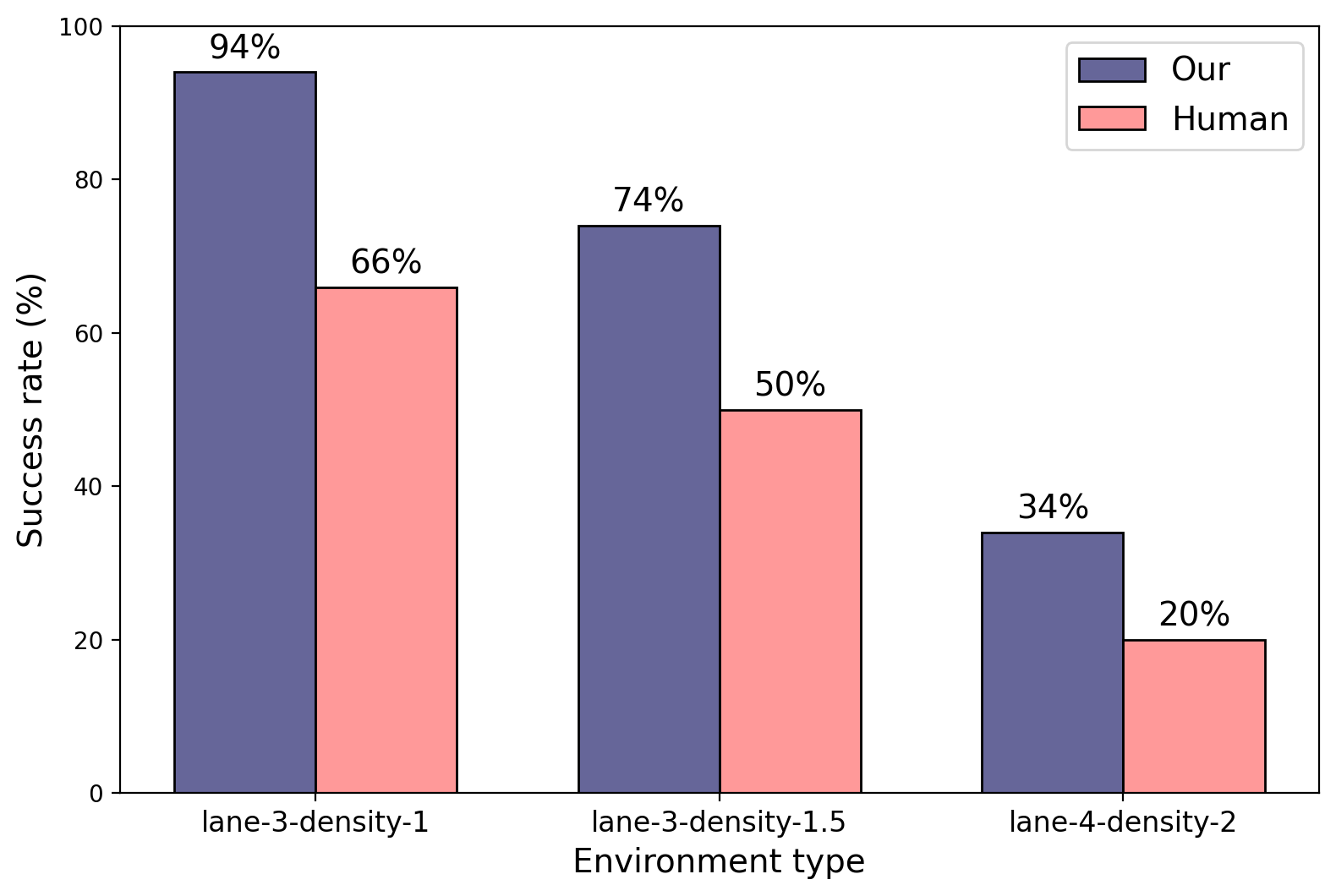}
\caption{Success rate comparison with human-designed reward in different types of highway environments.}
\label{Fig:success_rate}
\end{figure}

\begin{figure}
\centering
\includegraphics[width=3.4in]{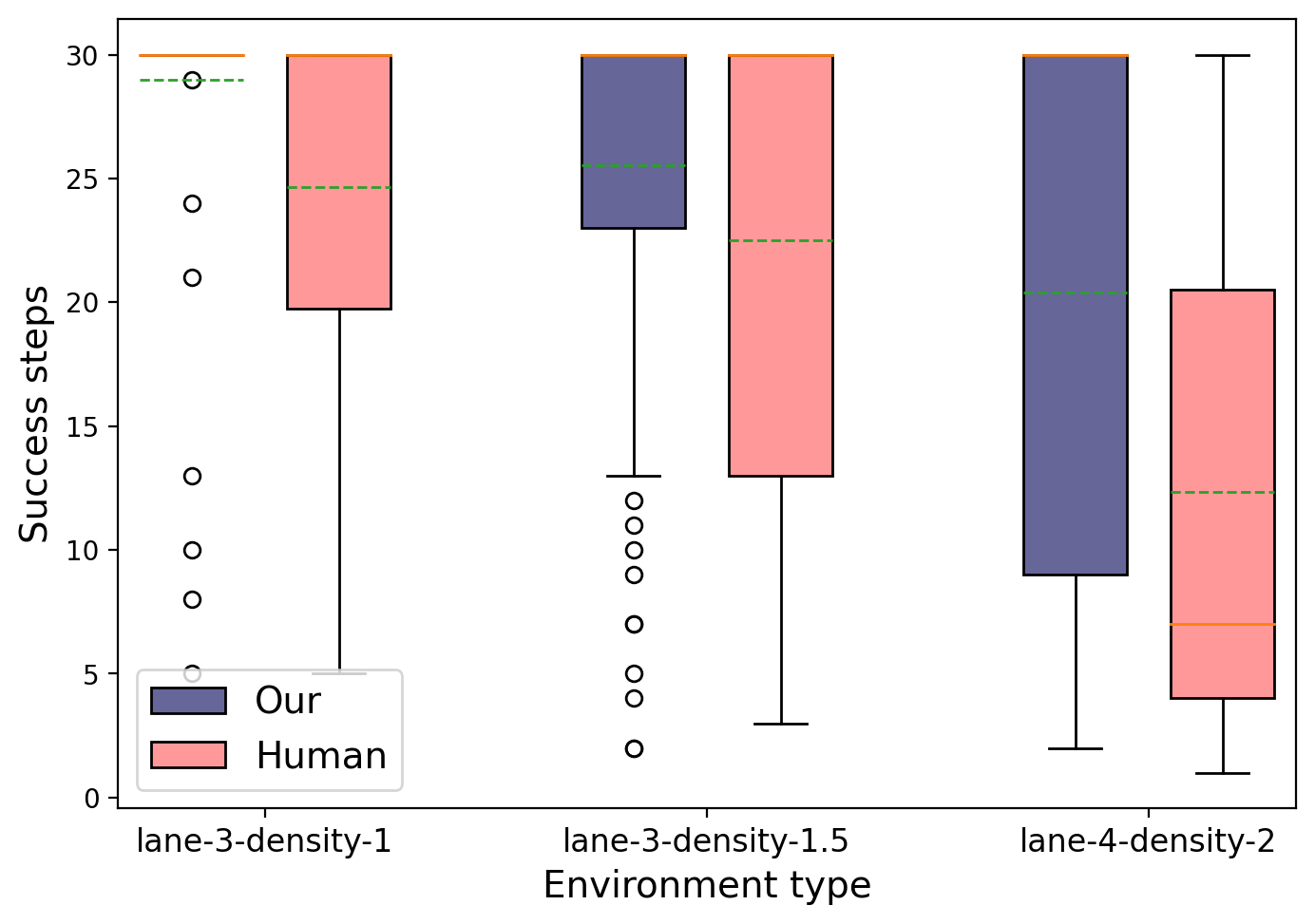}
\caption{Success steps comparison with human-designed reward in different types of highway environments.}
\label{Fig:speed}
\end{figure}

\subsection{Results}

We tested our reward design framework in three distinct environmental settings with varying complexities: \texttt{lane-3-density-1}, \texttt{lane-3-density-1.5}, and \texttt{lane-4-density-2}. A traffic flow density of 1 represents a relatively simple low-density scenario, while a density of 2 indicates a high-density scenario, where there is plenty of interaction between vehicles, making collisions more likely. Each environmental setting comprised 100 different scenarios with unique seeds. We assessed the performance of our reward designs by their success rates and success steps. A successful scenario is defined as the absence of collisions within 40 decision frames and the success step corresponds to the number of safe decision frames in the scenario. Despite the increased complexity in scenarios with higher vehicle densities and more lanes, as shown in Fig. \ref{Fig:success_rate} and Fig. \ref{Fig:speed}, our method consistently outperformed the human-designed rewards. On average, the success rate was 22\% higher across the various settings. Moreover, our framework emphasized better generalization ability than human reward across different scenarios according to the distributions of success steps. These results demonstrate that our system can generate reward functions that reliably enhance performance in various highway driving cases, significantly surpassing expert human-designed rewards.

\section{CONCLUSION} \label{section:conclusion}

Our research demonstrates a pioneering approach to reward function design in autonomous driving through the integration of LLMs and RL. By leveraging well designed prompt template and an iterative refinement process, our framework successfully generated high-quality, effective reward functions that significantly improved the success rate of highway driving scenarios in simulation. The results from extensive testing highlight a 22\% average increase in success rate over expert human-designed rewards, emphasizing enhanced safety in diverse driving scenarios. These findings suggest a promising direction for reducing the manual effort involved in reward function design and point towards the potential for LLMs to contribute significantly to the evolution of autonomous driving technologies.

Potential improvements for this study include using advanced prompting techniques such as Chain of Thought and Retrieval Augmented Generation to enhance the design of rewards with LLMs, and testing this method in more complex driving scenarios, such as at intersections and roundabouts.




{\appendices
\section*{Appendix} \label{section:appendix}

\subsection{Initial Prompt}

\begin{tcolorbox}
As an experienced engineer specializing in autonomous driving, your task is to design a reward function for the <Highway-env> simulator. The function will be written in Python and should effectively guide an autonomous vehicle in a simulated multilane highway environment.

Specifically, your objective is to generate a reward function that will help the ego-vehicle complete a reinforcement learning task. In this task, the ego-vehicle is driving on a multilane highway populated with other vehicles. The agent's objective is to reach a high speed while avoiding collisions with neighbouring vehicles. Driving on the right side of the road is also rewarded.

You will be given access to the Highway-env simulator's source code. While designing the reward function, pay close attention to the traffic configurations, such as the number of lanes and traffic density. These elements are crucial as they influence the interactions between the ego-vehicle and other vehicles on the road. Make necessary adjustments to your design to maximize its effectiveness.

Please ensure that your reward function complies with the following guidelines:
...

In the Highway-env Python code, please be aware of the relationship among the key components, particularly the HighwayEnv class, the ObservationType class, and their attributes, including Road and Vehicle. Here is the environment code:
...
\end{tcolorbox}

\subsection{Prompt for Reflection}

\begin{tcolorbox}
As an experienced engineer specializing in autonomous driving, your task is to improve a reward design for reinforcement learning with a highway driving task. In this task, the ego-vehicle is driving on a multilane highway populated with other vehicles. The agent's objective is to reach a high speed while avoiding collisions with neighbouring vehicles. Driving on the right side of the road is also rewarded. Based on the current reward function written in Python, reinforcement learning training has produced the following results in the agent:
...

Please carefully analyze the training results and create an improved Python reward function for the same task. If the results are unsatisfactory, consider redesigning it entirely. Otherwise, specific optimizations to certain parts of the reward function may be considered. Here is the current reward function:
...
\end{tcolorbox}
}

\bibliographystyle{IEEEtran}
\bibliography{my}

\end{document}